\documentclass[11pt]{amsart}

\calclayout

\usepackage[utf8]{inputenc}
\usepackage{fontenc}
\usepackage{amsfonts}
\usepackage{amssymb}
\usepackage{amsmath,pict2e}
\usepackage{amsthm}\usepackage{mathtools}
\usepackage{enumerate}
\usepackage{hyperref}\usepackage{enumitem}
\usepackage{mathrsfs}
\usepackage{tikz}
\usepackage{pgfplots}
\usepackage{marginnote}
\usepackage{xcolor,enumitem}
\usepackage{soul}
\usetikzlibrary{positioning,calc,fit,backgrounds}
\usepackage[all,cmtip]{xy}
 
\usepackage{tikz-cd}
\usepackage{float}
\usetikzlibrary{matrix,arrows,backgrounds,decorations.pathmorphing}
\newlength{\myarrowsize}
 
\pgfarrowsdeclare{cmto}{cmto}{
	\pgfsetdash{}{0pt} 
	\pgfsetbeveljoin 
	\pgfsetroundcap 
	\setlength{\myarrowsize}{0.6pt}
	\addtolength{\myarrowsize}{.5\pgflinewidth}
	\pgfarrowsleftextend{-4\myarrowsize-.5\pgflinewidth} 
	\pgfarrowsrightextend{.8\pgflinewidth}
}{
	\setlength{\myarrowsize}{0.6pt} 
  	\addtolength{\myarrowsize}{.5\pgflinewidth}  
	\pgfsetlinewidth{0.5\pgflinewidth}
	\pgfsetroundjoin
	\pgfpathmoveto{\pgfpoint{1.5\pgflinewidth}{0}}
	\pgfpatharc{-109}{-170}{4\myarrowsize}
	\pgfpatharc{10}{189}{0.58\pgflinewidth and 0.2\pgflinewidth}
	\pgfpatharc{-170}{-115}{4\myarrowsize+\pgflinewidth}
	\pgfpathclose
	\pgfusepathqfillstroke
	\pgfpathmoveto{\pgfpoint{1.5\pgflinewidth}{0}}
	\pgfpatharc{109}{170}{4\myarrowsize}
	\pgfpatharc{-10}{-189}{0.58\pgflinewidth and 0.2\pgflinewidth}
	\pgfpatharc{170}{115}{4\myarrowsize+\pgflinewidth}
	\pgfpathclose
	\pgfusepathqfillstroke
	\pgfsetlinewidth{2\pgflinewidth}
}

\pgfarrowsdeclare{cmonto}{cmonto}{
	\pgfsetdash{}{0pt} 
	\pgfsetbeveljoin 
	\pgfsetroundcap 
	\setlength{\myarrowsize}{0.6pt}
	\addtolength{\myarrowsize}{.5\pgflinewidth}
	\pgfarrowsleftextend{-4\myarrowsize-.5\pgflinewidth} 
	\pgfarrowsrightextend{.8\pgflinewidth}
}{
	\setlength{\myarrowsize}{0.6pt} 
  	\addtolength{\myarrowsize}{.5\pgflinewidth}  
	\pgfsetlinewidth{0.5\pgflinewidth}
	\pgfsetroundjoin
	\pgfpathmoveto{\pgfpoint{1.5\pgflinewidth}{0}}
	\pgfpatharc{-109}{-170}{4\myarrowsize}
	\pgfpatharc{10}{189}{0.58\pgflinewidth and 0.2\pgflinewidth}
	\pgfpatharc{-170}{-115}{4\myarrowsize+\pgflinewidth}
	\pgfpathclose
	\pgfusepathqfillstroke
	\pgfpathmoveto{\pgfpoint{1.5\pgflinewidth}{0}}
	\pgfpatharc{109}{170}{4\myarrowsize}
	\pgfpatharc{-10}{-189}{0.58\pgflinewidth and 0.2\pgflinewidth}
	\pgfpatharc{170}{115}{4\myarrowsize+\pgflinewidth}
	\pgfpathclose
	\pgfusepathqfillstroke
	\pgfpathmoveto{\pgfpoint{1.5\pgflinewidth-0.3em}{0}}
	\pgfpatharc{-109}{-170}{4\myarrowsize}
	\pgfpatharc{10}{189}{0.58\pgflinewidth and 0.2\pgflinewidth}
	\pgfpatharc{-170}{-115}{4\myarrowsize+\pgflinewidth}
	\pgfpathclose
	\pgfusepathqfillstroke
	\pgfpathmoveto{\pgfpoint{1.5\pgflinewidth-0.3em}{0}}
	\pgfpatharc{109}{170}{4\myarrowsize}
	\pgfpatharc{-10}{-189}{0.58\pgflinewidth and 0.2\pgflinewidth}
	\pgfpatharc{170}{115}{4\myarrowsize+\pgflinewidth}
	\pgfpathclose
	\pgfusepathqfillstroke
	\pgfsetlinewidth{2\pgflinewidth}
}

\pgfarrowsdeclare{cmhook}{cmhook}{
	\pgfsetdash{}{0pt} 
	\pgfsetbeveljoin 
	\pgfsetroundcap 
	\setlength{\myarrowsize}{0.6pt}
	\addtolength{\myarrowsize}{.5\pgflinewidth}
	\pgfarrowsleftextend{-4\myarrowsize-.5\pgflinewidth} 
	\pgfarrowsrightextend{.8\pgflinewidth}
}{
	\setlength{\myarrowsize}{0.6pt} 
  	\addtolength{\myarrowsize}{.5\pgflinewidth}  
 	\pgfsetdash{}{0pt}
	\pgfsetroundcap
	\pgfpathmoveto{\pgfqpoint{0pt}{-4.667\pgflinewidth}}
	\pgfpathcurveto
    {\pgfqpoint{4\pgflinewidth}{-4.667\pgflinewidth}}
    {\pgfqpoint{4\pgflinewidth}{0pt}}
    {\pgfpointorigin}
	\pgfusepathqstroke
}
\theoremstyle{Case1}

\theoremstyle{Case2}

\newtheorem*{rigprob*}{Rigidity Problem for uniform Roe Algebras}
\newtheorem*{rigprobcorona*}{Rigidity Problem for uniform Roe Coronas}

\newtheorem*{theorem*}{Theorem}

\newtheorem*{proposition*}{Proposition}

\newtheorem*{lemma*}{Lemma}

\newtheorem*{corollary*}{Corollar}

\newtheorem*{fact*}{Fact}

\theoremstyle{definition}

\newtheorem*{definition*}{Definition}

\newtheorem*{example*}{Example}

\newtheorem*{claim*}{Claim}

\newtheorem*{conjecture*}{Conjecture}

\theoremstyle{remark}

\newtheorem*{remark*}{Remark}

\newtheorem*{note*}{Note}
\newtheorem*{question*}{Question}

\makeatletter
\newcommand{\bigcomp}{%
  \DOTSB
  \mathop{\vphantom{\sum}\mathpalette\bigcomp@\relax}%
  \slimits@
}
\newcommand{\bigcomp@}[2]{%
  \begingroup\m@th
  \sbox\z@{$#1\sum$}%
  \setlength{\unitlength}{0.9\dimexpr\ht\z@+\dp\z@}%
  \vcenter{\hbox{%
    \begin{picture}(1,1)
    \bigcomp@linethickness{#1}
    \put(0.5,0.5){\circle{1}}
    \end{picture}%
  }}%
  \endgroup
}
\newcommand{\bigcomp@linethickness}[1]{%
  \linethickness{%
      \ifx#1\displaystyle 2\fontdimen8\textfont\else
      \ifx#1\textstyle 1.65\fontdimen8\textfont\else
      \ifx#1\scriptstyle 1.65\fontdimen8\scriptfont\else
      1.65\fontdimen8\scriptscriptfont\fi\fi\fi 3
  }%
}
\makeatother

\newcounter{my_enumerate_counter}
\newcommand{\pushcounter}{\setcounter{my_enumerate_counter}{\value{enumi}}}
\newcommand{\popcounter}{\setcounter{enumi}{\value{my_enumerate_counter}}}

\usepackage{enumitem}

\title{InAttention: Linear Context Scaling for Transformers}

\author[Joseph D. Eisner]{Joseph D. Eisner}

\begin{document}
\maketitle

\begin{abstract}
VRAM requirements for transformer models scale quadratically with context length due to the self-attention mechanism. In this paper we modify the decoder-only transformer, replacing self-attention with InAttention, which scales linearly with context length during inference by having tokens attend only to initial states. Benchmarking shows that InAttention significantly reduces VRAM usage during inference, enabling handling of long sequences on consumer GPUs. We corroborate that fine-tuning extends context length efficiently, improving performance on long sequences without high training costs. InAttention offers a scalable solution for long-range dependencies in transformer models, paving the way for further optimization.
\end{abstract}

%\textbf{Sparse Attention:}
%\cite{child2019generating} Local Attention
%\cite{zaheer2021big} Big Bird
%\cite{beltagy2020longformer} Longformer
%\cite{wang2020linformer} Linformer
%\cite{choromanski2022rethinking} Performer
%\cite{kitaev2020reformer} Reformer
%\cite{ren2021combiner} Combiner

%\textbf{Poor Context Length Extrapolation:}
%\cite{press2022train} Train Short
%\cite{sun2022lengthextrapolatable} Length Extrapolatable

%\textbf{Finetune Extend Context Length:} 
%\cite{chen2023extending}
%\cite{chen2024longlora} LongLora
%\cite{tworkowski2023focused} Focused Transformer
%\cite{mohtashami2023landmark} Landmark Attention

\section{Background and Introduction}
Decoder-based transformer stacks \cite{vaswani2023attention} have demonstrated syntactic and semantic understanding of language and other time-series data, achieving state-of-the-art few-shot \cite{brown2020language} performance across nearly any natural language task. They exhibit predictable scaling laws with respect to number of parameters and the amount of training data they ingest: bigger is better and we generally know by how much \cite{hoffmann2022training}. 

These \textit{Foundation Models} seem to benefit from extended context length but the self-attention mechanism at the heart of the transformer scales quadratically with context length -- making training and inference on extremely long queries cost prohibitive. One way to make long queries cheaper is to make the attention mechanism sparse, not allowing tokens to attend every token before them, but merely a subset.

Mohtashami and Jaggi \cite{mohtashami2023landmark} give an excellent summary\footnote{Citations updated to match our bibliography.} of sparse attention techniques: \begin{quote}
...For example,
Child et al. \cite{child2019generating} limit the attention to a local window around each token, while BigBird additionally
suggests attending to a random subset of previous tokens as well as several globally accessible
tokens \cite{zaheer2021big}. Longformer \cite{beltagy2020longformer} further introduces dilated sliding window patterns to increase attention’s
receptive field and manually picks the window sizes for each layer. Linformer \cite{wang2020linformer} uses a low-rank
approximation of the attention matrix while Performer \cite{choromanski2022rethinking} uses a non-softmax kernel to obtain a
more efficient implementation. Reformer \cite{kitaev2020reformer} uses locality-sensitive-hashing (LSH) to retrieve the
closest key vectors which should account for the highest scores in the attention matrix. Combiner \cite{ren2021combiner}
utilizes a hierarchical attention mechanism and heuristic reduction techniques, such as max-pooling,
to derive key and query vectors for input blocks...
\end{quote}

while their own approach involves inserting special \textit{landmark} tokens to stand in for consecutive blocks -- allowing models to restrict their attention to blocks which contain at least one high scoring token.

\begin{figure}[h]
\begin{tikzpicture}[auto, node distance=1cm,>=latex']
    \node (v00) {1};
    \node [right of=v00] (v10) {1};
    \node [right of=v10] (v20) {1};
    \node [right of=v20] (v30) {1};
    \node [right of=v30] (v40) {1};
    \node [right of=v40] (v50) {1};
    
    \node [above of=v00] (v01) {1};
    \node [right of=v01] (v11) {1};
    \node [right of=v11] (v21) {1};
    \node [right of=v21] (v31) {1};
    \node [right of=v31] (v41) {1};
    \node [right of=v41] (v51) {0};

    \node [above of=v01] (v02) {1};
    \node [right of=v02] (v12) {1};
    \node [right of=v12] (v22) {1};
    \node [right of=v22] (v32) {1};
    \node [right of=v32] (v42) {0};
    \node [right of=v42] (v52) {0};

    \node [above of=v02] (v03) {1};
    \node [right of=v03] (v13) {1};
    \node [right of=v13] (v23) {1};
    \node [right of=v23] (v33) {0};
    \node [right of=v33] (v43) {0};
    \node [right of=v43] (v53) {0};

    \node [above of=v03] (v04) {1};
    \node [right of=v04] (v14) {1};
    \node [right of=v14] (v24) {0};
    \node [right of=v24] (v34) {0};
    \node [right of=v34] (v44) {0};
    \node [right of=v44] (v54) {0};

    \node [above of=v04] (v05) {1};
    \node [right of=v05] (v15) {0};
    \node [right of=v15] (v25) {0};
    \node [right of=v25] (v35) {0};
    \node [right of=v35] (v45) {0};
    \node [right of=v45] (v55) {0};
    \node[fit=(v00)(v50)(v05)(v55), draw, ultra thick, rounded corners] (bdbox) {};

    \node at (2.5,-0.75) {\textit{Dense Attention Mask}};
\end{tikzpicture}
\hspace{2cm}
\begin{tikzpicture}[auto, node distance=1cm,>=latex']
    \node (v00) {0};
    \node [right of=v00] (v10) {0};
    \node [right of=v10] (v20) {0};
    \node [right of=v20] (v30) {0};
    \node [right of=v30] (v40) {1};
    \node [right of=v40] (v50) {1};
    
    \node [above of=v00] (v01) {0};
    \node [right of=v01] (v11) {0};
    \node [right of=v11] (v21) {0};
    \node [right of=v21] (v31) {1};
    \node [right of=v31] (v41) {1};
    \node [right of=v41] (v51) {0};

    \node [above of=v01] (v02) {0};
    \node [right of=v02] (v12) {0};
    \node [right of=v12] (v22) {1};
    \node [right of=v22] (v32) {1};
    \node [right of=v32] (v42) {0};
    \node [right of=v42] (v52) {0};

    \node [above of=v02] (v03) {0};
    \node [right of=v03] (v13) {1};
    \node [right of=v13] (v23) {1};
    \node [right of=v23] (v33) {0};
    \node [right of=v33] (v43) {0};
    \node [right of=v43] (v53) {0};

    \node [above of=v03] (v04) {1};
    \node [right of=v04] (v14) {1};
    \node [right of=v14] (v24) {0};
    \node [right of=v24] (v34) {0};
    \node [right of=v34] (v44) {0};
    \node [right of=v44] (v54) {0};

    \node [above of=v04] (v05) {1};
    \node [right of=v05] (v15) {0};
    \node [right of=v15] (v25) {0};
    \node [right of=v25] (v35) {0};
    \node [right of=v35] (v45) {0};
    \node [right of=v45] (v55) {0};

    \draw [-, red] (v40) -- (v41);
    \draw [-, red] (v41) -- (v31);
    \draw [-, red] (v31) -- (v32);
    \draw [-, red] (v32) -- (v22);
    \draw [-, red] (v22) -- (v23);
    \node[fit=(v00)(v50)(v05)(v55), draw, ultra thick, rounded corners] (bdbox) {};
    \node at (2.5,-0.75) {\textit{Sliding Window Mask}};
\end{tikzpicture}
\caption{Right: A sliding window mask with literal context length of 1. Outlined in red is the path information might take in a 3-layer transformer model giving an effective context length of 3.}
\label{fig:AttentionMasks}
\end{figure}
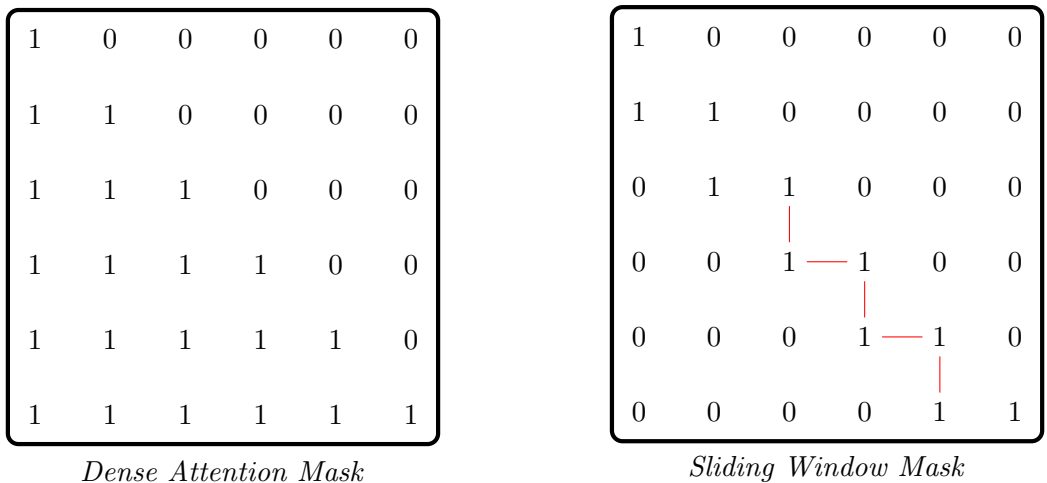

Perhaps the most human-interpretable form of sparse attention is \textit{sliding window} attention (e.g. Mistral's \cite{MistralAI}), which uses a lower-diagonal banded matrix as the attention mask. This bifurcates the notion of ``context length'': there is the \textit{literal context}, the length of the band, and there is an \textit{effective context} which is the length of the band multiplied by the number of layers, see Figure \ref{fig:AttentionMasks}. It is difficult to assess how well the effective context is truly utilized by the model.

In this paper we propose a technique for which \textbf{dense} attention scales linearly with context length. We do this by having the hidden states at each layer attend the initial states of a query, rather than themselves, inspiring the name \textit{InAttention}. This has two major downstream effects on inference:

\begin{enumerate}
    \item The attention matrix becomes an attention vector, since we do not need to calculate any hidden states except the ultimate one. For long queries, the attention matrix represents an enormous overhead, potentially dwarfing the size of the model being used. Consequently, we dramatically reduce the maximum VRAM needed for the initial next-token prediction.
    \item After the initial next-token prediction, it is common practice to cache the hidden states at each layer so they do not need to be recomputed. This prevents us from needing to spin up the attention matrix again but it still represents substantial overhead. With InAttention, since we only attend initial states, we do not need to do this -- freeing up VRAM even downstream of the initial surge from (1).
\end{enumerate}

Our experiments suggest that there is a small but real decrease in model capability (measured by training loss) by using InAttention, which we will describe in Section \ref{Capability}. However, we consider this a small price to pay for the gains and suggest the capability be bolstered by allocating some of the freed-up VRAM to additional model parameters, see Figure \ref{fig:LossVsVRAM}.

While InAttention allows for inference on extremely long queries, it does not make it any cheaper to train the model, where we will still want to compute losses for every token prediction in a training batch and so the cost is still quadratic. To partially address this, we corroborate that models can be cheaply finetuned to extend their context length (as suggested in e.g. \cite{chen2023extending}, \cite{tworkowski2023focused}, and \cite{chen2024longlora}), our analysis is in Section \ref{Finetuning}.

Our hope is that further efficiencies will be found which, paired with InAttention, will allow for both cheaper training and inference on long queries. While InAttention on its own does not solve training, it does essentially solve inference, and we hope this is a major step towards running exceedingly long queries on consumer hardware.

\subsection{Why Compute Scales Quadratically With Context Length}\label{WhyQuadratic}

We refer to the sequence of vectors connected via residual connections as a \textit{residual tower}. Our perspective is that this sequence represents a vector ``morphing'' or ``mutating'' as it ascends the transformer stack.

% We want him to be in the maximum uncertainty, so that his mind will be filled with contradictory pictures of the future, every one of which arouses hope or fear. There is nothing like suspense and anxiety for barricading a human’s mind against the Enemy. He wants men to be concerned with what they do; our business is to keep them thinking about what will happen to them. -Screwtape Letters (Letter 6)

During the self-attention step of the transformer each token $t$ must attend a substantial number of other tokens $\sigma(t)$, depending on the mask, and this number grows linearly as the number of tokens $T$ is increased, $\sigma(t) \sim T$. So the total number of attention pairs we must compute is $\sum_t \sigma(t) \sim \sum_t T \sim T^2$.

In the case of decoder-based transformers, which use a causal mask, each token must attend itself and every token which comes before it, yielding $\frac{T^2 + T}{2}$ attention pairs in each self-attention mechanism  given a context length of $T$ tokens.

Sparse attention addresses quadratic scaling by bounding $\sigma(t) < C$, $C \in \mathbb{N}$, so the number of attention computations becomes $\sum_t \sigma(t) \leq \sum_t C = CT$, thus achieving linear scaling with context length. In the next subsection we will share some intuition for InAttention but a mathematically simple summary is this: Where sparse attention sees the expression $\sum_t \sigma(t)$ and seeks to bound it by controlling $\sigma$, InAttention instead looks to remove the sum.

\section{InAttention: Attending Initial Tokens Only}\label{InAttention}
\subsection{Intuition for InAttention}\label{Intuition}
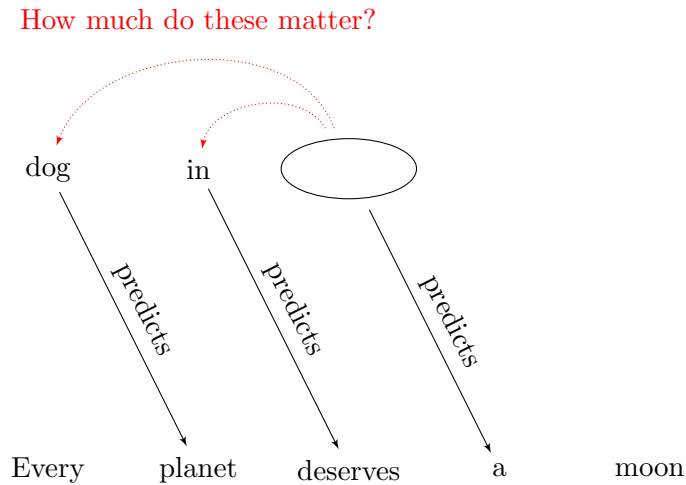
\begin{figure}[h]
\begin{tikzpicture}[auto, node distance=2cm,>=latex']
    \node (Every) {Every};
    \node [right of=Every] (planet) {planet};
    \node [right of=planet] (deserves) {deserves};
    \node [right of=deserves] (a) {a};
    \node [right of=a] (moon) {moon};

    \node [above of=Every] (blank) {};

    \node [above of=blank] (dog) {dog};
    \node [right of=dog] (in) {in};
    \node [right of=in] (circle) {
    \begin{tikzpicture}
      \draw (0,0) ellipse (0.9 and 0.4);
    \end{tikzpicture}
    };

    \draw [->] (dog) -- (planet) node [midway, above, sloped] (TextNode) {predicts};
    \draw [->] (in) -- (deserves) node [midway, above, sloped] (TextNode) {predicts};
    \draw [->] (circle) -- (a) node [midway, above, sloped] (TextNode) {predicts};

    \draw [densely dotted, red, ->] (circle) to [out=120,in=80,loop,looseness=1.0] (in);
    \draw [densely dotted, red, ->] (circle) to [out=110,in=70,loop,looseness=1.0] (dog) ;
    
    \node [above of=in, red] (attend) {How much do these matter?};
    
\end{tikzpicture}
\caption{Next Token Game Demonstration}
\label{fig:NextTokenGame}
\end{figure}

When the token-vectors work their way to the ``top'' of the transformer stack, each represents the model's latent approximation of a prediction for the next token. For instance, given the phrase ``Every planet deserves a moon'', the vector originally representing ``Every'' morphs into (a latent approximation of) the models prediction for the token following ``Every'', perhaps ``dog'', anticipating ``Every dog has its day''. The vector originally representing ``planet'' likewise morphs into the model's prediction for the token following ``planet'', perhaps ``in'' anticipating ``Every planet in the solar system''. It is interesting that in predicting the token following ``planet'' we have the vector attending the (typically) false prediction ``dog'' (or a latent approximation thereof). Reference Figure \ref{fig:NextTokenGame}.

It is conceivable (and appears to be the case, see Section \ref{Capability}
) that the latent representation which predicts ``dog'' also contains some partial work towards subsequent tokens, which the model can utilize. But it also seems plausible that many of the features in that prediction are not useful, and much of the work is immediately depreciated with the revealing of the true next token, ``planet''.

During inference, the final layer of vectors is mostly uninteresting, since the linear decoder layer does not contain a self-attention mechanism. The aforementioned predictions ``dog'' and ``in'' are discarded -- only the vector in the final position will be decoded for generation. Thus the penultimate layer can be dramatically simplified: instead of $\frac{T^2 + T}{2}$ attention pairs we only need to compute attention for the vector in the final position -- resulting in precisely $T$ attention pairs. All of the other attention pairs add nothing, as they are used to predict tokens which we already know. There appears to be a hint buried here: perhaps we could similarly discard the corresponding states in earlier layers? This would leave us only having to compute the residual tower sitting above the ultimate token.

We summarize our observations in Figure \ref{fig:AttentionPairs}, showing a 3-layer transformer stack:

\tikzstyle{block} = [draw, fill=white, rectangle, rounded corners,
    minimum height=6em, minimum width=3em]
\tikzstyle{hblock} = [draw, fill=white, rectangle, rounded corners, 
    minimum height=3em, minimum width=21em]
\tikzstyle{sum} = [draw, fill=white, circle, node distance=1cm]
\tikzstyle{input} = [coordinate]
\tikzstyle{output} = [coordinate]
\tikzstyle{pinstyle} = [pin edge={to-,thin,black}]
\tikzset{scribble/.style={decorate, decoration={zigzag, amplitude=.4cm, segment length=.415cm, aspect=.75}},}

\begin{figure}[h]
\scalebox{0.60}{
\begin{tikzpicture}[auto, node distance=2cm,>=latex']

    \node [hblock, name=encoder] {Input Embedding};
    \node[block, above of=encoder, fill=cyan!20] (v_03) {$v_3$};
    \node [block, left of=v_03, fill=green!20] (v_02) {$v_2$};
    \node [block, left of=v_02, fill=yellow!20] (v_01) {$v_1$};
    \node [block, right of=v_03, fill=orange!20] (v_04) {$v_4$};
    \node [block, right of=v_04, fill=purple!20] (v_05) {$v_5$};
    \node [hblock, above of=v_03] (transformer1) {Transformer Layer};

    \node [right=0.5cm of v_05] {$15$ attention pairs};

    \node [block, above of=transformer1, fill=cyan!20] (v_13) {};
    \node [block, left of=v_13, fill=green!20] (v_12) {};
    \node [block, left of=v_12, fill=yellow!20] (v_11) {};
    \node [block, right of=v_13, fill=orange!20] (v_14) {};
    \node [block, right of=v_14, fill=purple!20] (v_15) {};
    \node [hblock, above of=v_13] (transformer2) {Transformer Layer};

    \node [right=0.5cm of v_15] {$15$ attention pairs};

    \node [block, above of=transformer2, fill=cyan!20] (v_23) {};
    \node [block, left of=v_23, fill=green!20] (v_22) {};
    \node [block, left of=v_22, fill=yellow!20] (v_21) {};
    \node [block, right of=v_23, fill=orange!20] (v_24) {};
    \node [block, right of=v_24, fill=purple!20] (v_25) {};
    \node [hblock, above of=v_23] (transformer3) {Transformer Layer};

    \node [right=0.5cm of v_25,align=center] {$15$ attention pairs\\only $5$ are needed\\(those in purple)};

    \node [block, above of=transformer3, fill=cyan!20] (v_33) {};
    \node [block, left of=v_33, fill=green!20] (v_32) {};
    \node [block, left of=v_32, fill=yellow!20] (v_31) {};
    \node [block, right of=v_33, fill=orange!20] (v_34) {};
    \node [block, right of=v_34, fill=purple!20] (v_35) {};
    \node [hblock, above of=v_33] (decoder) {Linear Decoder};
    \begin{scope}[on background layer]
    %%% Attention Pairs for layer 0
    \draw [->, color=purple!60!black] (v_05.118) -- (v_01.62);
    \draw [->, color=purple!60!black] (v_05.120) -- (v_02.60);
    \draw [->, color=purple!60!black] (v_05.123) -- (v_03.57);
    \draw [->, color=purple!60!black] (v_05.126) -- (v_04.54);
    \draw [->, color=purple!60!black] (v_05.290) to [out=330,in=350,loop,looseness=2.4] (v_05.330);

    \draw [->, color=orange!60!black] (v_04.128) -- (v_01.52);
    \draw [->, color=orange!60!black] (v_04.132) -- (v_02.48);
    \draw [->, color=orange!60!black] (v_04.136) -- (v_03.44);
    \draw [->, color=orange!60!black] (v_04.290) to [out=330,in=350,loop,looseness=2.4] (v_04.330);

    \draw [->, color=cyan!60!black] (v_03.139) -- (v_01.41);
    \draw [->, color=cyan!60!black] (v_03.144) -- (v_02.36);
    \draw [->, color=cyan!60!black] (v_03.290) to [out=330,in=350,loop,looseness=2.4] (v_03.330);

    \draw [->, color=green!60!black] (v_02.148) -- (v_01.32);
    \draw [->, color=green!60!black] (v_02.290) to [out=330,in=350,loop,looseness=2.4] (v_02.330);

    \draw [->, color=yellow!60!black] (v_01.290) to [out=330,in=350,loop,looseness=2.4] (v_01.330);

    %%% Attention Pairs for layer 1
    \draw [->, color=purple!60!black] (v_15.118) -- (v_11.62);
    \draw [->, color=purple!60!black] (v_15.120) -- (v_12.60);
    \draw [->, color=purple!60!black] (v_15.123) -- (v_13.57);
    \draw [->, color=purple!60!black] (v_15.126) -- (v_14.54);
    \draw [->, color=purple!60!black] (v_15.290) to [out=330,in=350,loop,looseness=2.4] (v_15.330);

    \draw [->, color=orange!60!black] (v_14.128) -- (v_11.52);
    \draw [->, color=orange!60!black] (v_14.132) -- (v_12.48);
    \draw [->, color=orange!60!black] (v_14.136) -- (v_13.44);
    \draw [->, color=orange!60!black] (v_14.290) to [out=330,in=350,loop,looseness=2.4] (v_14.330);

    \draw [->, color=cyan!60!black] (v_13.139) -- (v_11.41);
    \draw [->, color=cyan!60!black] (v_13.144) -- (v_12.36);
    \draw [->, color=cyan!60!black] (v_13.290) to [out=330,in=350,loop,looseness=2.4] (v_13.330);

    \draw [->, color=green!60!black] (v_12.148) -- (v_11.32);
    \draw [->, color=green!60!black] (v_12.290) to [out=330,in=350,loop,looseness=2.4] (v_12.330);

    \draw [->, color=yellow!60!black] (v_11.290) to [out=330,in=350,loop,looseness=2.4] (v_11.330);

    %%% Attention Pairs for layer 2
    \draw [->, color=purple!60!black] (v_25.118) -- (v_21.62);
    \draw [->, color=purple!60!black] (v_25.120) -- (v_22.60);
    \draw [->, color=purple!60!black] (v_25.123) -- (v_23.57);
    \draw [->, color=purple!60!black] (v_25.126) -- (v_24.54);
    \draw [->, color=purple!60!black] (v_25.290) to [out=330,in=350,loop,looseness=2.4] (v_25.330);

    \draw [->, color=orange!60!black] (v_24.128) -- (v_21.52);
    \draw [->, color=orange!60!black] (v_24.132) -- (v_22.48);
    \draw [->, color=orange!60!black] (v_24.136) -- (v_23.44);
    \draw [->, color=orange!60!black] (v_24.290) to [out=330,in=350,loop,looseness=2.4] (v_24.330);

    \draw [->, color=cyan!60!black] (v_23.139) -- (v_21.41);
    \draw [->, color=cyan!60!black] (v_23.144) -- (v_22.36);
    \draw [->, color=cyan!60!black] (v_23.290) to [out=330,in=350,loop,looseness=2.4] (v_23.330);

    \draw [->, color=green!60!black] (v_22.148) -- (v_21.32);
    \draw [->, color=green!60!black] (v_22.290) to [out=330,in=350,loop,looseness=2.4] (v_22.330);

    \draw [->, color=yellow!60!black] (v_21.290) to [out=330,in=350,loop,looseness=2.4] (v_21.330);
    \end{scope}
    \draw [scribble, red] (v_31.85) -- (v_31.270);
    \draw [scribble, red] (v_32.90) -- (v_32.275);
    \draw [scribble, red] (v_33.85) -- (v_33.275);
    \draw [scribble, red] (v_34.95) -- (v_34.265);
\end{tikzpicture}
\quad
\begin{tikzpicture}[auto, node distance=2cm,>=latex']
    \node [hblock, name=encoder] {Input Embedding};
    \node[block, above of=encoder, fill=cyan!20] (v_03) {$v_3$};
    \node [block, left of=v_03, fill=green!20] (v_02) {$v_2$};
    \node [block, left of=v_02, fill=yellow!20] (v_01) {$v_1$};
    \node [block, right of=v_03, fill=orange!20] (v_04) {$v_4$};
    \node [block, right of=v_04, fill=purple!20] (v_05) {$v_5$};
    \node [hblock, above of=v_03, opacity=0.8, text opacity=1] (transformer1) {Transformer Layer};

    \node [right=0.5cm of v_05] {$5$ attention pairs};

    \node [above of=transformer1] (v_13) {};
    \node [left of=v_13] (v_12) {};
    \node [left of=v_12] (v_11) {};
    \node [right of=v_13] (v_14) {};
    \node [block, right of=v_14, fill=purple!20] (v_15) {};
    \node [hblock, above of=v_13, opacity=0.8, text opacity=1] (transformer2) {Transformer Layer};

    \node [right=0.5cm of v_15] {$5$ attention pairs};

    \node [above of=transformer2] (v_23) {};
    \node [left of=v_23] (v_22) {};
    \node [left of=v_22] (v_21) {};
    \node [right of=v_23] (v_24) {};
    \node [block, right of=v_24, fill=purple!20] (v_25) {};
    \node [hblock, above of=v_23] (transformer3) {Transformer Layer};

    \node [right=0.5cm of v_25] {$5$ attention pairs};

    \node [above of=transformer3] (v_33) {};
    \node [left of=v_33] (v_32) {};
    \node [left of=v_32] (v_31) {};
    \node [right of=v_33] (v_34) {};
    \node [block, right of=v_34, fill=purple!20] (v_35) {};
    \node [hblock, above of=v_33] (decoder) {Linear Decoder};
    \begin{scope}[on background layer]
    %%% Attention Pairs for layer 0
    \draw [->, color = purple!80!black] (v_05.118) -- (v_01.62);
    \draw [->, color = purple!80!black] (v_05.120) -- (v_02.60);
    \draw [->, color = purple!80!black] (v_05.123) -- (v_03.57);
    \draw [->, color = purple!80!black] (v_05.126) -- (v_04.54);
    \draw [->, color = purple!80!black] (v_05.290) to [out=330,in=350,loop,looseness=2.4] (v_05.330);

    %%% Attention Pairs for layer 1
    \draw [->, color = purple!80!black] (v_15.210) -- (v_01.67);
    \draw [->, color = purple!80!black] (v_15.230) -- (v_02.68);
    \draw [->, color = purple!80!black] (v_15.250) -- (v_03.69);
    \draw [->, color = purple!80!black] (v_15.270) -- (v_04.70);
    \draw [->, color = purple!80!black] (v_15.290) -- (v_05.80);

    %%% Attention Pairs for layer 2
    \draw [->, color = purple!80!black] (v_25.180) -- (v_01.90);
    \draw [->, color = purple!80!black] (v_25.200) -- (v_02.88);
    \draw [->, color = purple!80!black] (v_25.220) -- (v_03.86);
    \draw [->, color = purple!80!black] (v_25.240) -- (v_04.84);
    \draw [->, color = purple!80!black] (v_25.310) to [out=310,in=30,loop,looseness=0.25] (v_05.30);
    \end{scope}

\end{tikzpicture}
}
\caption{Left: Visualized attention pairs for dense attention. Right: Visualized attention pairs for inattention.}
\label{fig:AttentionPairs}
\end{figure}
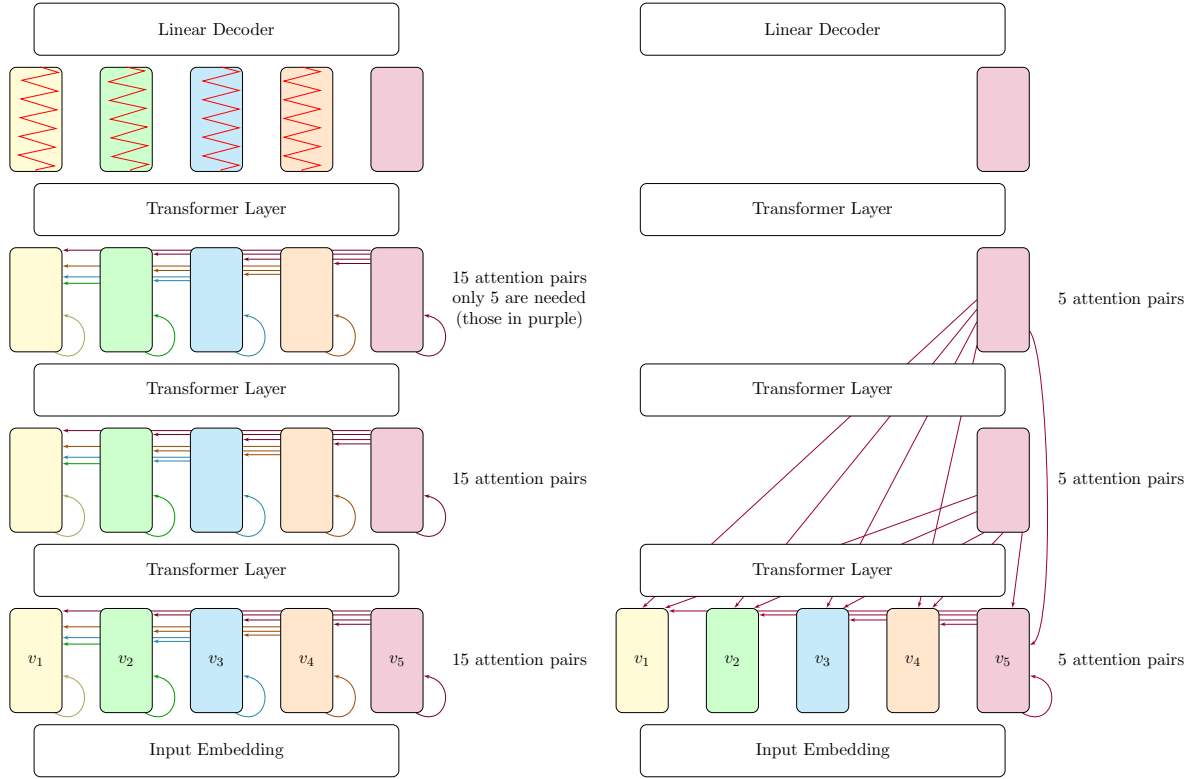

\subsection{InAttention Definition}\label{InAttentionEquations}
We recall a few of the observations we made in Section \ref{Intuition}:
\begin{enumerate}
    \item Each residual tower allocates some (perhaps much) of its work and features to predicting the immediate next token, work which is irrelevant for residual towers to the right and invisible for those to the left.
    \item Work and features a residual tower ``pays forward'' to help predict future tokens is, by its nature, speculative and immediately depreciates in value when the actual next token is revealed.
    \item The penultimate layer of a transformer stack safely ignores all attention-pairs except those involving the final position. This means the final layer could safely have $T$ (number of input tokens) attention pairs instead of the $\frac{T^2 + T}{2}$ of other layers.
\end{enumerate}

Given these observations, we recommend the following modification to the transformer stack: Have tokens attend initial states instead of states in their own row. During inference this means the vast majority of states need not be computed (only the ultimate residual tower), and the number of attention pairs goes down dramatically, and scales linearly. These changes are illustrated in Figure \ref{fig:AttentionPairs}. 

In some sense we are removing ``self-attention'' where columns in the tensor of hidden states attend each other, instead they attend the columns in the tensor of initial states. In another sense, however, this might be considered a variation of self-attention, where (a latent representation) of language tokens are attending (a latent representation) of themselves -- we are merely changing where the latent representations are coming from.

Figure \ref{fig:CircuitDiagrams}    \footnote{Tikz code adapted from \cite{MiltosKofinas}.} provides circuit diagrams for GPTNeoX models with and without our InAttention adjustment:
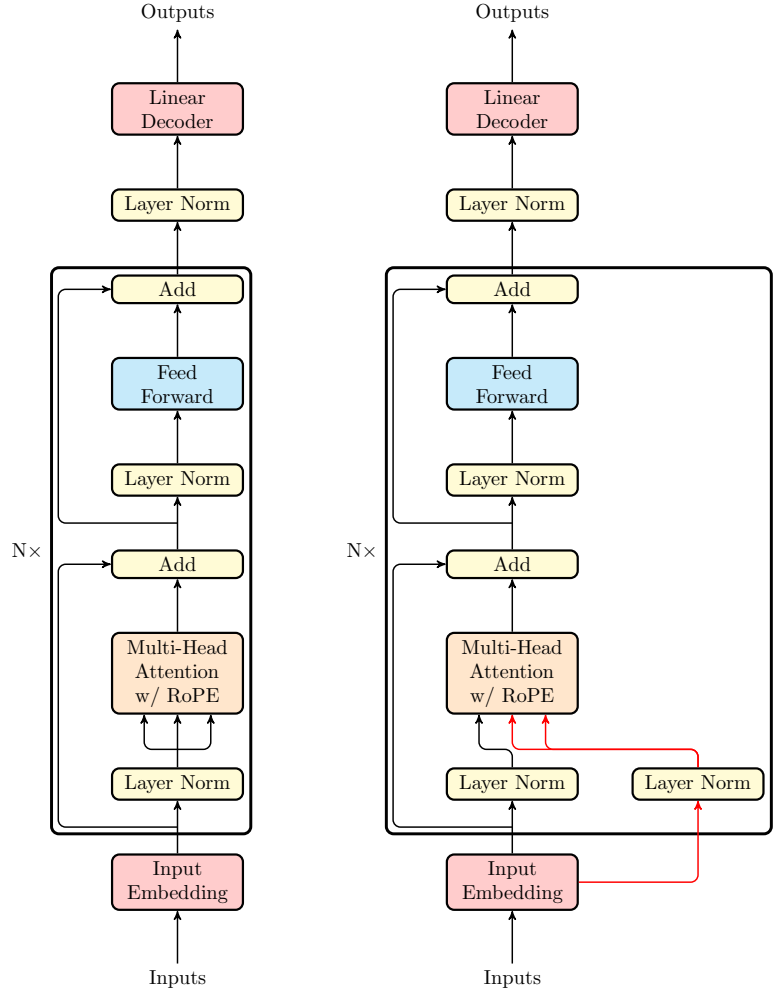
\begin{figure}[H]
\scalebox{.70}{
\begin{tikzpicture}[
        module/.style={draw, very thick, rounded corners, minimum width=15ex},
        embmodule/.style={module, fill=red!20},
        mhamodule/.style={module, fill=orange!20},
        lnmodule/.style={module, fill=yellow!20},
        ffnmodule/.style={module, fill=cyan!20},
        arrow/.style={-stealth', thick, rounded corners},
        ]       
    \node (inputs) {Inputs};
    \node[above=of inputs, embmodule, align=center]
    (inputemb) {Input\\Embedding};
    \node[above=of inputemb, lnmodule, align=center] (norm1)
    {Layer Norm};

    \node[above=of norm1, mhamodule, align=center] (mha)
    {Multi-Head\\Attention\\w/ RoPE};
    \node[above=of mha, lnmodule, align=center] (add1)
    {Add};
    \node[above=of add1, lnmodule, align=center] (norm2)
    {Layer Norm};
    \node[above=of norm2, ffnmodule, align=center] (ffn)
    {Feed\\Forward};
    \node[above=of ffn, lnmodule, align=center] (add2)
    {Add};
    \node[above=of add2, lnmodule, align=center] (normf)
    {Layer Norm};
    \node[above=of normf, embmodule, align=center] (embedout) {Linear\\Decoder};
    \node[above=of embedout] (outputs) {Outputs};

    \draw[arrow] (inputs) -- (inputemb);
    \draw[arrow] (inputemb) -- (norm1);
    \draw[arrow] (mha) -- (add1);
    \draw[arrow] (add1) -- (norm2);
    \draw[arrow] (norm2) -- (ffn);
    \draw[arrow] (ffn) -- (add2);
    \draw[arrow] (add2) -- (normf);
    \draw[arrow] (normf) -- (embedout);
    \draw[arrow] (embedout) -- (outputs);

    \coordinate (mharesidual) at ($(inputemb.north)!0.5!(norm1.south)$);
    \coordinate (ffnresidual) at ($(norm2.south)!0.5!(add1.north)$);
    \coordinate (mhafork) at ($(mha.south)!0.5!(norm1)$);
    \coordinate[left=of add1] (ln1residualleft);
    \coordinate[left=of add2] (ln2residualleft);

    \node[fit=(norm1)(mha)(add1)(add2)(mharesidual)(ln1residualleft), draw, ultra thick, rounded corners, label=left:$\mathrm{N\times}$] (encoder) {};
    \draw[arrow] (mharesidual)-|(ln1residualleft)--(add1);
    \draw[arrow] (ffnresidual)-|(ln2residualleft)--(add2);
    \draw[arrow] (mhafork)-|($(mha.south)!0.5!(mha.south west)$);
    \draw[arrow] (mhafork)-|($(mha.south)!0.5!(mha.south east)$);

    \begin{scope}[on background layer]
        \draw[arrow] (norm1)--(mha);
    \end{scope}
\end{tikzpicture}\hspace{4em}
\begin{tikzpicture}[
        module/.style={draw, very thick, rounded corners, minimum width=15ex},
        embmodule/.style={module, fill=red!20},
        mhamodule/.style={module, fill=orange!20},
        lnmodule/.style={module, fill=yellow!20},
        ffnmodule/.style={module, fill=cyan!20},
        arrow/.style={-stealth', thick, rounded corners},
        ]       
    \node (inputs) {Inputs};
    \node[above=of inputs, embmodule, align=center]
    (inputemb) {Input\\Embedding};
    \node[above=of inputemb, lnmodule, align=center] (norm1)
    {Layer Norm};
    \node[above right=of inputemb, lnmodule, align=center] (normn)
    {Layer Norm};

    \node[above=of norm1, mhamodule, align=center] (mha)
    {Multi-Head\\Attention\\w/ RoPE};
    \node[above=of mha, lnmodule, align=center] (add1)
    {Add};
    \node[above=of add1, lnmodule, align=center] (norm2)
    {Layer Norm};
    \node[above=of norm2, ffnmodule, align=center] (ffn)
    {Feed\\Forward};
    \node[above=of ffn, lnmodule, align=center] (add2)
    {Add};
    \node[above=of add2, lnmodule, align=center] (normf)
    {Layer Norm};
    \node[above=of normf, embmodule, align=center] (embedout) {Linear\\Decoder};
    \node[above=of embedout] (outputs) {Outputs};

    \draw[arrow] (inputs) -- (inputemb);
    \draw[arrow] (inputemb) -- (norm1);
    \draw[arrow] (mha) -- (add1);
    \draw[arrow] (add1) -- (norm2);
    \draw[arrow] (norm2) -- (ffn);
    \draw[arrow] (ffn) -- (add2);
    \draw[arrow] (add2) -- (normf);
    \draw[arrow] (normf) -- (embedout);
    \draw[arrow] (embedout) -- (outputs);

    \coordinate (mharesidual) at ($(inputemb.north)!0.5!(norm1.south)$);
    \coordinate (ffnresidual) at ($(norm2.south)!0.5!(add1.north)$);
    \coordinate (mhafork) at ($(mha.south)!0.5!(norm1)$);
    \coordinate[above=of normn] (peright1);
    \coordinate[above=of peright1] (peright2);
    \coordinate (valueinput) at ($(mha.south)!0.5!(mha.south east)$);
    \coordinate[right=of mhafork] (initsplit);
    \coordinate[left=of add1] (ln1residualleft);
    \coordinate[left=of add2] (ln2residualleft);

    \node[fit=(norm1)(normn)(mha)(add1)(add2)(mharesidual)(ln1residualleft), draw, ultra thick, rounded corners, label=left:$\mathrm{N\times}$] (encoder) {};
    \draw[arrow] (mharesidual)-|(ln1residualleft)--(add1);
    \draw[arrow] (ffnresidual)-|(ln2residualleft)--(add2);
    \draw[arrow, red] (normn)|-(initsplit)-|(valueinput);
    \draw[arrow, red] (inputemb)-|(normn);
    \begin{scope}[on background layer]
        \draw[arrow] (norm1)--(mhafork)-|($(mha.south)!0.5!(mha.south west)$);
        \draw[arrow, red] (normn)|-(initsplit)-|(mha);
    \end{scope}
\end{tikzpicture}
}
\caption{Left: GPTNeoX Model (without parallel residuals). Right: GPTNeoX Model modified to use InAttention. Red edges indicate the path of initial states.}
\label{fig:CircuitDiagrams}
\end{figure}

Scaled dot-product attention (equation (1) from \cite{vaswani2023attention}) can be represented as \[ \textrm{Attention}(Q, K, V) = \textrm{softmax}\left(\frac{QK^T}{\sqrt{d_k}}\right)V \] In a decoder-only transformer stack we have that $Q = W_Q X; K = W_K X; V = W_V X$ (assuming no biases). Here $W_Q, W_K$ and $W_V$ are linear operators and $X$ represents the hidden-states of the model being fed in. 

InAttention is very similar but instead $Q = W_Q X; K = W_K Y$ and $V = W_V Y$ where $Y$ represents the initial-states of the model, which come straight from the Input Embedding and pass through a layer-specific Layer Norm.

\section{Benchmarking Inattention}
In this section we will quantify the effect that Inattention has on VRAM footprint and model capability. We will be training a series of baseline models using the $GPTNeoX$ architecture. We bring our own \textit{Nazareth} tokenizer, which aims to be a word-level tokenizer specializing in English words. We refer to this series of baseline models as the $NazX$ series.

Modifying $GPTNeoX$ to implement InAttention is straightforward. We track the initial states and pass them to each layer. Attention queries are still computed by applying a linear transform $W_Q$ to the hidden states, while the attention keys and values are now obtained by applying linear transforms $W_K$ and $W_V$ to the initial states. The only other architectural difference is that we initialize an additional layernorm for each attention block, which is applied to the initial states. This results in our $NazXN$, InAttention based model series having slightly more parameters than their $NazX$ counterparts.

Table \ref{fig:NazXHypers} summarizes the hyper-parameters\footnote{All models use 8 attention heads.} of the $NazX$ and $NazXN$ model series:
\begin{figure}[h]
\begin{tabular}[hbt!]{ |p{2.5cm}|p{2.5cm}|p{2.5cm}|p{2.5cm}|p{2.5cm}|  }
 \hline
 \multicolumn{5}{|c|}{Models} \\
 \hline
 Model Name &Embed Dim &MLP Factor &Layers &Parameters\\
 \hline
 $NazX235$ & 0768 & 3 & 12 & 235610880 \\
 $NazX420$ & 1024 & 4 & 16 & 421168128 \\
 $NazX735$ & 1280 & 5 & 20 & 733646080 \\
 $NazXN235$ & 0768 & 3 & 12 & 235629312 \\
 $NazXN420$ & 1024 & 4 & 16 & 421200896 \\
 $NazXN735$ & 1280 & 5 & 20 & 733697280 \\
 \hline
\end{tabular}
\caption{NazX(N) Family of Models}
\label{fig:NazXHypers}
\end{figure}

\subsection{VRAM Footprint}
The promise of Inattention is a substantially smaller VRAM footprint during inference. To illustrate this, we run queries of various lengths on $NazX420$ and $NazXN420$ architectures, keeping track of the VRAM usage and yielding the following results:
\\\\
\begin{figure}[h]
\begin{tabular}[hbt!]{ |p{4cm}|p{4cm}|p{4cm}| }
 \hline
 \multicolumn{3}{|c|}{VRAM Usage} \\
 \hline
 Query Length (Tokens) &NaxX VRAM (MB) &NazXN VRAM (MB) \\
 \hline
 Model Loaded & 2945 & 2869 \\
 01024 & 4675 & 2965 \\
 02048 & 8271 & 2967 \\
 04096 & 18831 & 3231 \\
 08192 & 58339 & 4337 \\
 16384 & OOM & 6503 \\
 32768 & OOM & 10883 \\
 \hline
\end{tabular}
\caption{420M Series VRAM Comparison}
\label{fig:420VRAMChart}
\end{figure}

VRAM usage now scales linearly with context length and gains are substantial even for relatively short queries. Changing the model size will primarily affect the y-intercepts, so graphs would appear similar for various model sizes and architectures.

Notably, query lengths which cannot squeeze within an NVIDIA H100 graphics card for a typical transformer model (regardless of parameter count) fit comfortably on a 16GB VRAM GPU, not uncommon among consumers, using Inattention.

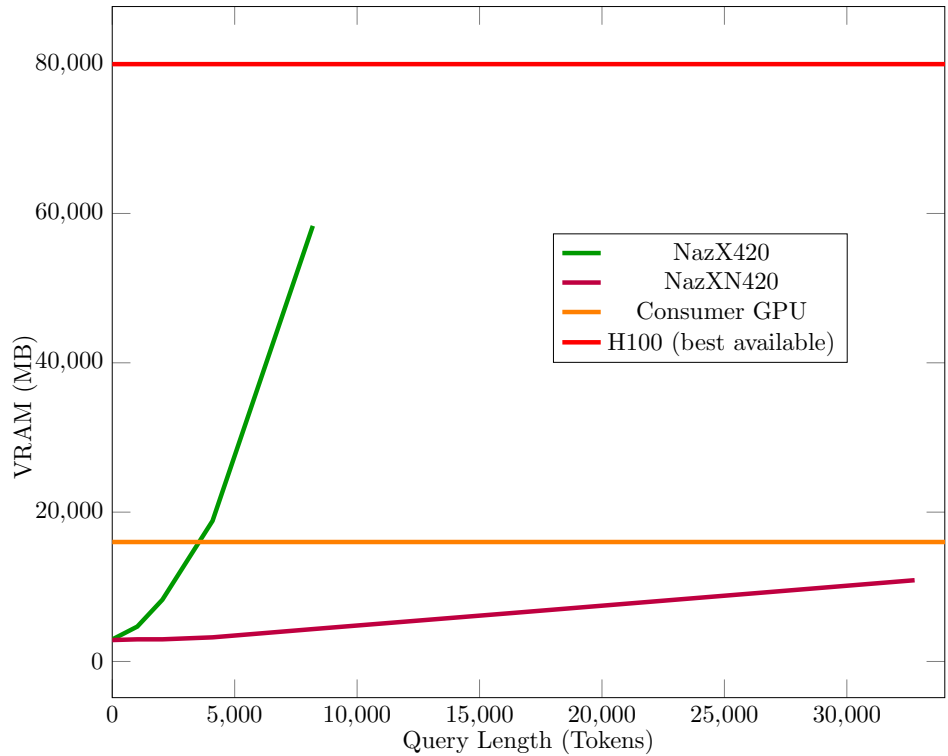
\begin{figure}[H]
\scalebox{.85}{
\begin{tikzpicture}
  \begin{axis}[xlabel=Query Length (Tokens), ylabel=VRAM (MB), domain=0:36000, xmin=0, xmax=34000, xscale=1.9, yscale=1.9, yticklabel style={/pgf/number format/fixed}, xticklabel style={/pgf/number format/fixed}, scaled y ticks = false, scaled x ticks = false, legend style={at={(axis cs:18000,40000)},anchor=south west}, xlabel style = {at={(axis cs:17000,-3500)},anchor=north}, ylabel style = {at={(axis cs:-350,45000)},anchor=east},]
    \addplot[green!60!black, line width=2pt] coordinates {
      (0, 2945)
      (1024,  4675)
      (2048,  8271)
      (4096,  18831)
      (8192, 58339)
    };
    \addlegendentry{NazX420}
    \addplot[purple, line width=2pt] coordinates {
      (0, 2869)
      (1024,  2965)
      (2048,  2967)
      (4096,  3231)
      (8192, 4337)
      (16384, 6503)
      (32768, 10883)
    };
    \addlegendentry{NazXN420}
    \addplot[mark=none, orange, line width=2pt, samples=2] {16000};
    \addlegendentry{Consumer GPU}
    \addplot[mark=none, red, line width=2pt, samples=2] {80000};
    \addlegendentry{H100 (best available)}
  \end{axis}
\end{tikzpicture}
}
\caption{420M Series VRAM Comparison}
\label{fig:420VRAMGraph}
\end{figure}

\subsection{Capability}\label{Capability}
Of course running on a small amount of VRAM means nothing if Inattention-based models do not exhibit similar intelligence and scaling as their Attention-based counterparts, so we explore capabilities next.

In the following results, NazX and NazXN models were trained on the first $235, 420$ or $735$ $C4$ files, respectively. We were unable to tokenize file 23 of $C4$, so it is omitted from all ranges and the subsequent file included instead (for instance, NazX420 was trained on files $[0, 22] \cup [24, 420]$).

Models were trained on a cluster of 16 $P4DE$ instances on AWS, each consisting of 8 A100 GPUs with 80GB VRAM. We used Deepspeed, leveraging the Hugging Face Trainer with default AdamW optimizer and cosine annealing scheduler with an initial learning rate of $2*10^{-4}$. 

We adjusted batches per device for each model to roughly optimize VRAM usage but adjusted gradient accumulation steps in tandem so that the total number of forward steps (the product of batches per device, number of devices (which for us was $16*8=128$), and gradient accumulation steps) per optimizer step was $GPUS*BS*GAS = 12288$. The $235M$ and $420M$ models were trained with a batch size of $12$, while the $735M$ models were trained with a batch size of $8$.

We then run evaluations on file $1000$ of $C4$, tokenized at different max context lengths, and report the average loss over this file in all upcoming figures.

\begin{figure}[H]
\scalebox{0.85}{
\begin{tikzpicture}
  \begin{axis}[xlabel=Query Length (Tokens), ylabel=Eval Loss, domain=32:5000, xmin=28, xmax=5000, xscale=1.8, yscale=1.8, yticklabel style={/pgf/number format/fixed}, xticklabel style={/pgf/number format/fixed}, scaled y ticks = false, scaled x ticks = false, legend style={at={(axis cs:512,4.0)},anchor=south west}, xlabel style = {at={(axis cs:512,2.8)},anchor=north}, ylabel style = {at={(axis cs:31,3.7)},anchor=south},xmode=log, log basis x={2}, log ticks with fixed point,]

    \addplot[purple, line width=2pt, dotted] coordinates {
      (32,4.403885841369629)
      (64,4.1478047370910645)
      (128,3.9647438526153564)
      (256,3.8448612689971924)
      (512,3.7735538482666016)
      (1024,3.7391083240509033)
      (2048,3.8345954418182373)
      (4096,3.8300364017486572)
    };
    \addlegendentry{NazXN235}
    \addplot[green!60!black, line width=2pt, dotted] coordinates {
      (32,4.350996494293213)
      (64,4.071300029754639)
      (128,3.8581316471099854)
      (256,3.7084925174713135)
      (512,3.6116912364959717)
      (1024,3.5613350868225098)
      (2048,3.635601043701172)
      (4096,3.648639678955078)
    };
    \addlegendentry{NazX235}
    \addplot[purple, line width=2pt, dashed, dash pattern=on 9pt off 3pt] coordinates {
      (32,4.122905731201172)
      (64,3.8338534832000732)
      (128,3.6227526664733887)
      (256,3.482391595840454)
      (512,3.397632360458374)
      (1024,3.355663299560547)
      (2048,3.371324062347412)
      (4096,3.3920071125030518)
    };
    \addlegendentry{NazXN420}
    \addplot[green!60!black, line width=2pt, dashed, dash pattern=on 9pt off 3pt] coordinates {
      (32,4.0656046867370605)
      (64,3.7568891048431396)
      (128,3.5249135494232178)
      (256,3.3629629611968994)
      (512,3.2584280967712402)
      (1024,3.2036244869232178)
      (2048,3.221599578857422)
      (4096,3.2491674423217773)
    };
    \addlegendentry{NazX420}
    \addplot[purple, line width=2pt] coordinates {
      (32,3.9180705547332764)
      (64,3.608211040496826)
      (128,3.3806650638580322)
      (256,3.2291951179504395)
      (512,3.139674186706543)
      (1024,3.105959177017212)
      (2048,3.183635711669922)
      (4096,3.235863447189331)
    };
    \addlegendentry{NazXN735}
    \addplot[green!60!black, line width=2pt] coordinates {
      (32,3.833815574645996)
      (64,3.5048649311065674)
      (128,3.261047840118408)
      (256,3.095428705215454)
      (512,2.9925127029418945)
      (1024,2.948413133621216)
      (2048,3.024336099624634)
      (4096,3.074554204940796)
    };
    \addlegendentry{NazX735}
  \end{axis}
\end{tikzpicture}
}
\caption{Loss By Context Length (Models Trained At CL=1024)}
\end{figure}
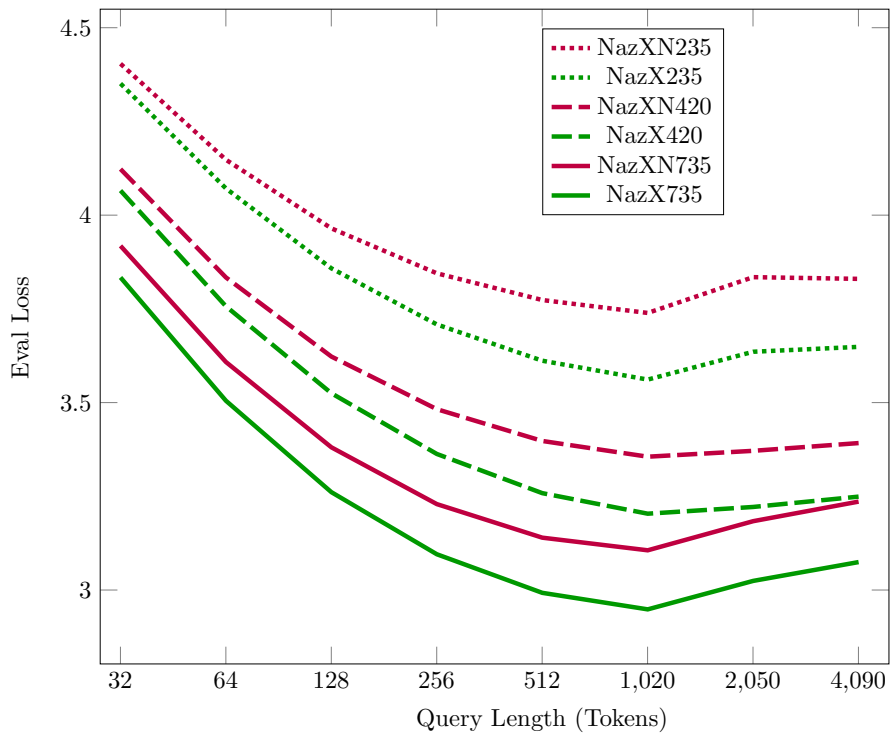
Capability degradation is non-negligible when switching to Inattention, but the tax does not appear to be growing with model size -- we consider the trade quite favorable. The freed VRAM can be leveraged for capability gain in other ways -- the most naive of which is to simply run a larger model.

Cutting the data another way, we below consider loss plotted against VRAM footprint for models running inference at a maximum context length of 1024 (which is what they were each pretrained with). While difficult to extrapolate from, the results for our six models are favorable towards InAttention and this advantage will rapidly increase as query lengths increase.

\begin{figure}[H]
\begin{tikzpicture}
  \begin{axis}[xlabel=Inference VRAM (MB), ylabel=Eval Loss, domain=1500:8000, xmin=1500, xmax=8000, xscale=1.8, yscale=1.8, yticklabel style={/pgf/number format/fixed}, xticklabel style={/pgf/number format/fixed}, scaled y ticks = false, scaled x ticks = false, legend style={at={(axis cs:5000,3.6)},anchor=south west}, xlabel style = {at={(axis cs:4750,2.875)},anchor=north}, ylabel style = {at={(axis cs:1650,3.35)},anchor=south}]

    \addplot[purple, line width=2pt] coordinates {
      (2077,3.7391083240509033)
      (2965,3.355663299560547)
      (4345,3.105959177017212)
    };
    \addlegendentry{InAttention Models}
    \addplot[green!60!black, line width=2pt] coordinates {
      (3677,3.5613350868225098)
      (4515,3.2036244869232178)
      (7295,2.948413133621216)
    };
    \addlegendentry{Dense Attention Models}
    
    \node[label={0:{NazXN235}},circle,fill,purple,inner sep=2pt] at (axis cs:2077,3.7391083240509033) {};
    \node[label={210:{NazXN420}},circle,fill,purple,inner sep=2pt] at (axis cs:2965,3.355663299560547) {};
    \node[label={225:{NazXN735}},circle,fill,purple,inner sep=2pt] at (axis cs:4345,3.1059591770172123) {};

    \node[label={0:{NazX235}},circle,fill,green!60!black,inner sep=2pt] at (axis cs:3677,3.5613350868225098) {};
    \node[label={45:{NazX420}},circle,fill,green!60!black,inner sep=2pt] at (axis cs:4515,3.2036244869232178) {};
    \node[label={260:{NazX735}},circle,fill,green!60!black,inner sep=2pt] at (axis cs:7295,2.948413133621216) {};
  \end{axis}
\end{tikzpicture}
\caption{Loss By VRAM Footprint (At CL=1024)}
\label{fig:LossVsVRAM}
\end{figure}
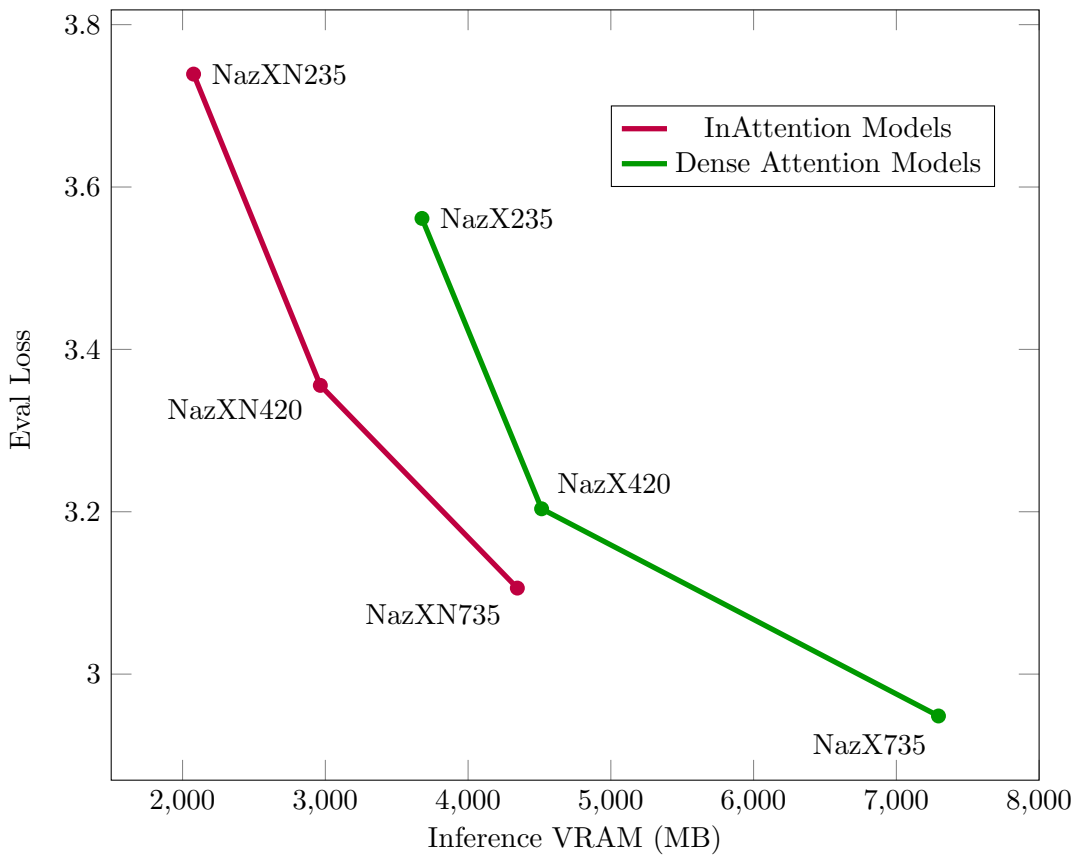

\section{Utilizing Long Context}
\begin{figure}
\begin{tikzpicture}
    \begin{axis}[
        domain=32:5000,
        xmin=28,
        xmax=5000,
        xscale=0.9,
        ymin=2.8,
        ymax=5.2,
        yscale=0.9,
        yticklabel style={/pgf/number format/fixed},
        xticklabel style={/pgf/number format/fixed},
        scaled y ticks=false,
        scaled x ticks=false,
        legend style={at={(axis cs:32,4.3)}, anchor=south west},
        ylabel style={at={(axis cs:48,4.0)}, anchor=south},
        xmode=log,
        log basis x={2},
        xlabel style={at={(axis cs:362,2.8)}, anchor=north,align=center, text width=5cm},
        xlabel={Query Length (Tokens) \\ Trained at CL=128},
        ylabel=Eval Loss,
    ]
    \addplot[purple, line width=2pt, dashed, dash pattern=on 9pt off 3pt] coordinates {
        (32,3.8194706439971924)
      (64,3.5498738288879395)
      (128,3.3610525131225586)
      (256,3.8836543560028076)
      (512,4.695240497589111)
      (1024,5.001677513122559)
      (2048,4.926947116851807)
      (4096,4.624871253967285)
    };
    \addlegendentry{NazXN420}
    \addplot[green!60!black, line width=2pt, dashed, dash pattern=on 9pt off 3pt] coordinates {
        (32,3.751068592071533)
      (64,3.462843179702759)
      (128,3.257143974304199)
      (256,3.5934619903564453)
      (512,4.322108268737793)
      (1024,4.630001068115234)
      (2048,4.5637898445129395)
      (4096,4.295501708984375)
    };
    \addlegendentry{NazX420}
    \end{axis}
\end{tikzpicture}
\begin{tikzpicture}
    \begin{axis}[
        domain=32:5000,
        xmin=28,
        xmax=5000,
        xscale=0.9,
        ymin=2.8,
        ymax=5.2,
        yscale=0.9,
        yticklabel style={/pgf/number format/fixed},
        xticklabel style={/pgf/number format/fixed},
        scaled y ticks=false,
        scaled x ticks=false,
        legend style={at={(axis cs:4090,4.3)}, anchor=south east},
        ylabel style={at={(axis cs:31,3.7)}, anchor=south},
        xmode=log,
        log basis x={2},
        xlabel style={at={(axis cs:362,2.8)}, anchor=north,align=center, text width=5cm},
        xlabel={Query Length (Tokens) \\ Trained at CL=1024},
    ]
    \addplot[purple, line width=2pt, dashed, dash pattern=on 9pt off 3pt] coordinates {
        (32,4.122905731201172)
        (64,3.8338534832000732)
        (128,3.6227526664733887)
        (256,3.482391595840454)
        (512,3.397632360458374)
        (1024,3.355663299560547)
        (2048,3.371324062347412)
        (4096,3.3920071125030518)
    };
    \addlegendentry{NazXN420}
    \addplot[green!60!black, line width=2pt, dashed, dash pattern=on 9pt off 3pt] coordinates {
        (32,4.0656046867370605)
        (64,3.7568891048431396)
        (128,3.5249135494232178)
        (256,3.3629629611968994)
        (512,3.2584280967712402)
        (1024,3.2036244869232178)
        (2048,3.221599578857422)
        (4096,3.2491674423217773)
    };
    \addlegendentry{NazX420}
    \end{axis}
\end{tikzpicture}
\caption{Length Extrapolation (Or Lack Thereof)}
\label{fig:ContextLengthExtrap}
\end{figure}
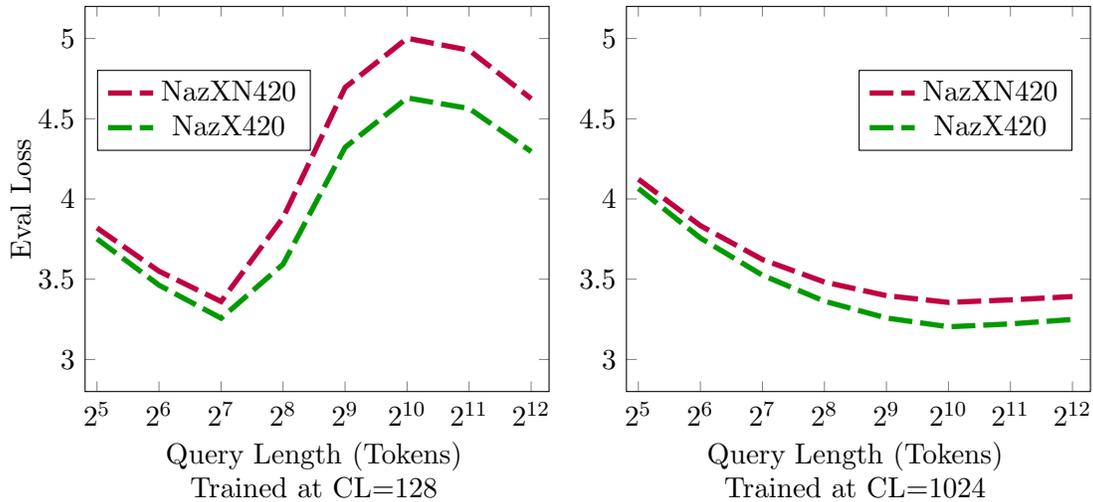

Inattention allows substantially cheaper inference on long queries, but does not buy us any efficiency during training. Capability degradation of models past their trained context length is well established (e.g. \cite{press2022train}, \cite{sun2022lengthextrapolatable}), but we illustrate the problem by pretraining a NazX and NazXN model at a context length of 128, and then run inference at various context lengths, noting substantial capbility degradation at unseen context lengths, see the left graph in Figure \ref{fig:ContextLengthExtrap}.

Models simply do not learn to extrapolate to unseen context lengths, despite our best attempts to provide positional encodings with built-in inductive biases. 

We want our models to use as long of a context as possible, to better leverage the inference efficiency of Inattention, but the cost of doing so is prohibitive. A context length of 8192 does not fit on our GPUs if we train naively.

\subsection{Finetuning On Long Context Queries}\label{Finetuning}
Luckily there is some evidence (e.g. \cite{chen2023extending}, \cite{tworkowski2023focused}, \cite{chen2024longlora}) that models can be cheaply fine-tuned to extend their context length. We seek to verify this independently: 

Using a NazX420 model pretrained at context length 128 as a starting point we finetune on a single additional file of $C4$, file 421, tokenized at a context length of 1024. We treat this as a new training job with freshly initialized AdamW optimizer and cosine anneal scheduler with an initial learning rate of $2*10^{-5}$, one order of magnitude smaller than our initial pretraining learning rate. We also run an identical fine-tuning regime on a NazXN420 model pretrained at a context length of 128, presenting the results for both below.
\begin{figure}[H]
\begin{tikzpicture}
    \begin{axis}[
        domain=32:5000,
        xmin=28,
        xmax=5000,
        xscale=0.9,
        ymin=2.8,
        ymax=5.2,
        yscale=0.9,
        yticklabel style={/pgf/number format/fixed},
        xticklabel style={/pgf/number format/fixed},
        scaled y ticks=false,
        scaled x ticks=false,
        legend style={at={(axis cs:32,4.3)}, anchor=south west},
        ylabel style={at={(axis cs:48,4.0)}, anchor=south},
        xmode=log,
        log basis x={2},
        xlabel style={at={(axis cs:362,2.8)}, anchor=north,align=center, text width=5cm},
        xlabel={Query Length (Tokens) \\ Trained at CL=128},
        ylabel=Eval Loss,
    ]
    \addplot[purple, line width=2pt, dashed, dash pattern=on 9pt off 3pt, opacity=0.5] coordinates {
      (32,3.8194706439971924)
      (64,3.5498738288879395)
      (128,3.3610525131225586)
      (256,3.8836543560028076)
      (512,4.695240497589111)
      (1024,5.001677513122559)
      (2048,4.926947116851807)
      (4096,4.624871253967285)
    };
    \addlegendentry{NazXN420}
    \addplot[green!60!black, line width=2pt, dashed, dash pattern=on 9pt off 3pt, opacity=0.5] coordinates {
      (32,3.751068592071533)
      (64,3.462843179702759)
      (128,3.257143974304199)
      (256,3.5934619903564453)
      (512,4.322108268737793)
      (1024,4.630001068115234)
      (2048,4.5637898445129395)
      (4096,4.295501708984375)
    };
    \addlegendentry{NazX420}
    \addplot[purple, line width=2pt] coordinates {
      (32,3.854330062866211)
      (64,3.5843777656555176)
      (128,3.395855665206909)
      (256,3.2960002422332764)
      (512,3.260798931121826)
      (1024,3.2519752979278564)
      (2048,3.250958204269409)
      (4096,3.2376699447631836)
    };
    \addlegendentry{NazXN420FT}
    \addplot[green!60!black, line width=2pt] coordinates {
      (32,3.751068592071533)
      (64,3.494335174560547)
      (128,3.2859694957733154)
      (256,3.160459518432617)
      (512,3.09586238861084)
      (1024,3.070605516433716)
      (2048,3.0680651664733887)
      (4096,3.0682907104492188)
    };
    \addlegendentry{NazX420FT}
    \legend{,,NazXN420FT,NazX420FT}
    \end{axis}
\end{tikzpicture}
\begin{tikzpicture}
    \begin{axis}[
        domain=32:5000,
        xmin=28,
        xmax=5000,
        xscale=0.9,
        ymin=2.8,
        ymax=5.2,
        yscale=0.9,
        yticklabel style={/pgf/number format/fixed},
        xticklabel style={/pgf/number format/fixed},
        scaled y ticks=false,
        scaled x ticks=false,
        legend style={at={(axis cs:4090,4.3)}, anchor=south east},
        ylabel style={at={(axis cs:31,3.7)}, anchor=south},
        xmode=log,
        log basis x={2},
        xlabel style={at={(axis cs:362,2.8)}, anchor=north,align=center, text width=5cm},
        xlabel={Query Length (Tokens) \\ Trained at CL=1024},
    ]
    \addplot[purple, line width=2pt, dashed, dash pattern=on 9pt off 3pt, opacity=0.5] coordinates {
        (32,4.122905731201172)
        (64,3.8338534832000732)
        (128,3.6227526664733887)
        (256,3.482391595840454)
        (512,3.397632360458374)
        (1024,3.355663299560547)
        (2048,3.371324062347412)
        (4096,3.3920071125030518)
    };
    \addlegendentry{NazXN420}
    \addplot[green!60!black, line width=2pt, dashed, dash pattern=on 9pt off 3pt, opacity=0.5] coordinates {
        (32,4.0656046867370605)
        (64,3.7568891048431396)
        (128,3.5249135494232178)
        (256,3.3629629611968994)
        (512,3.2584280967712402)
        (1024,3.2036244869232178)
        (2048,3.221599578857422)
        (4096,3.2491674423217773)
    };
    \addlegendentry{NazX420}
    \addplot[purple, line width=2pt] coordinates {
      (32,3.854330062866211)
      (64,3.5843777656555176)
      (128,3.395855665206909)
      (256,3.2960002422332764)
      (512,3.260798931121826)
      (1024,3.2519752979278564)
      (2048,3.250958204269409)
      (4096,3.2376699447631836)
    };
    \addlegendentry{NazXN420FT}
    \addplot[green!60!black, line width=2pt] coordinates {
      (32,3.751068592071533)
      (64,3.494335174560547)
      (128,3.2859694957733154)
      (256,3.160459518432617)
      (512,3.09586238861084)
      (1024,3.070605516433716)
      (2048,3.0680651664733887)
      (4096,3.0682907104492188)
    };
    \addlegendentry{NazX420FT}
    \legend{,,NazXN420FT,NazX420FT}
    \end{axis}
\end{tikzpicture}
\caption{Finetuning to Extend Context Length}
\end{figure}
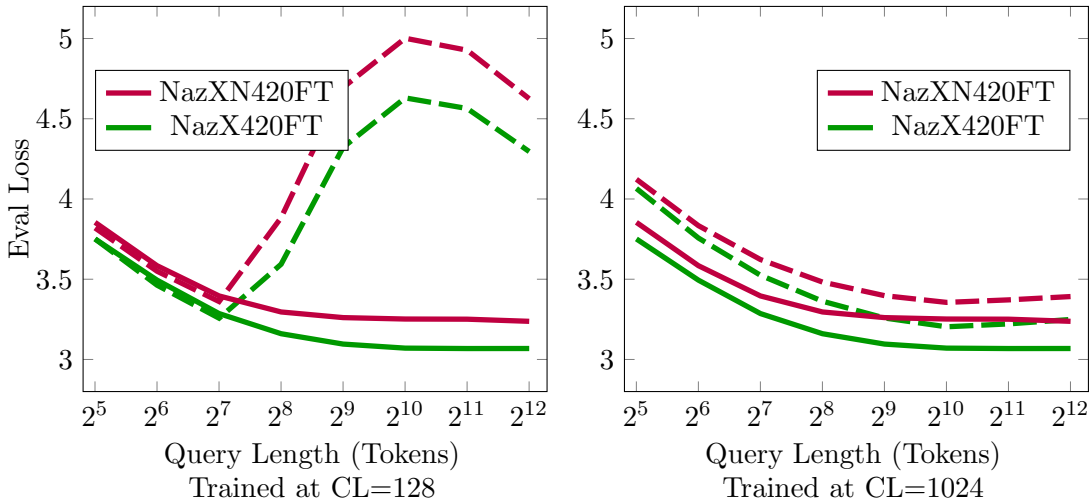

Indeed, finetuning on less than 1\% of the initial training data allows the the CL128 models to dramatically improve their performance at CL1024, with only marginal degradation in performance on short context queries.

There are three things to observe in the figures above:
\begin{enumerate}
    \item Finetuning on unseen context length data appears to be very effective at extending the context length a model knows how to use.
    \item Models finetuned in this way better generalize to still unseen context lengths -- note the slope is flat or slightly negative when increasing context length to the unseen lengths of $2^{11}$ and $2^{12}$. This is in contrast to merely pretrained models which show performance degradation past their trained context length almost immediately.
    \item The models pretrained at a context length of 128 and finetuned at 1024 outperform the models pretrained at context length 1024.
\end{enumerate}

We urge caution when interpreting item (3) above -- we suspect this is an artifact of our hyper-parameters: we did not increase the ratio of forward passes to optimizer steps for the models trained at a context length of 128, meaning those models received more optimizer steps than the models trained at 1024 context length (since the data was sharded into more, shorter batches).

The first two items, however, suggest a way forward: pretrain models at a reasonably long but still practical context length, then cheaply finetune them on much longer context lengths.

\section{Conclusion and Next Steps}\label{Conclusion}

In this paper, we introduced InAttention, a modified attention mechanism that enables linear scaling of compute and memory with respect to context length during transformer inference. By having the hidden states at each layer attend to the initial token embeddings rather than the previous layer's states, InAttention reduces the attention matrix to a vector and eliminates the need to cache intermediate activations.

Our experiments demonstrate that InAttention substantially reduces VRAM usage compared to standard dense attention, with the gains increasing for longer sequence lengths. This improved efficiency comes at the cost of model capability, as measured by evaluation loss, but this capability degradation can be accounted for by running larger models. We show that freed up memory from InAttention can be leveraged to deploy larger models that are more capable than smaller dense attention models with equivalent VRAM usage.

While InAttention provides a significant improvement in inference efficiency, it does not reduce the computational burden of training transformers with very long contexts. However, we corroborated recent findings showing that models can be cheaply finetuned to handle longer sequences than they were initially trained on, achieving strong performance. We believe the most promising path forward is to pretrain InAttention models with reasonably long contexts, then finetune them to fully exploit the inference benefits of InAttention on even longer sequences.

Future work should explore techniques to further optimize the efficiency of training InAttention-based models on long sequences. Combining InAttention with sparse attention mechanisms, more sophisticated finetuning methods, and other architectural innovations may lead to even greater gains. Overall, we believe InAttention is a valuable step towards practical and scalable transformer models that can fully leverage long-range context, and we are excited to see further developments in this direction.

\bibliographystyle{plain}
\bibliography{refs}
\end{document}